# A Physics-informed Deep Operator for Real-Time Freeway Traffic State Estimation

**Hongxin Yu**
College of Civil Engineering and Architecture
Zhejiang University, Hangzhou, China, 310058
Email: yuhongxin@zju.edu.cn

**Yibing Wang\***
College of Civil Engineering and Architecture
Zhejiang University, Hangzhou, China, 310058
Email: wangyibing@zju.edu.cn

**Fengyue Jin**
College of Civil Engineering and Architecture
Zhejiang University, Hangzhou, China, 310058
Email: jinfengyue@zju.edu.cn

**Meng Zhang**
College of Civil Engineering and Architecture
Zhejiang University, Hangzhou, China, 310058
Email: 12412177@zju.edu.cn

**Anni Chen**
College of Civil Engineering and Architecture
Zhejiang University, Hangzhou, China, 310058
Email: 22312301@zju.edu.cn

*Hongxin Yu, Yibing Wang, Fengyue Jin, Meng Zhang, and Anni Chen*

**ABSTRACT**

Traffic state estimation (TSE) falls methodologically into three categories: model-driven, data-driven, and model-data dual-driven. Model-driven TSE relies on macroscopic traffic flow models originated from hydrodynamics. Data-driven TSE leverages historical sensing data and employs statistical models or machine learning methods to infer traffic state. Model-data dual-driven traffic state estimation attempts to harness the strengths of both aspects to achieve more accurate TSE. From the perspective of mathematical operator theory, TSE can be viewed as a type of operator that maps available measurements of inerested traffic state into unmeasured traffic state variables in real time. For the first time this paper proposes to study real-time freeway TSE in the idea of physics-informed deep operator network (PI-DeepONet), which is an operator-oriented architecture embedding traffic flow models based on deep neural networks. The paper has developed an extended architecture from the original PI-DeepONet. The extended architecture is featured with: (1) the acceptance of 2-D data input so as to support CNN-based computations; (2) the introduction of a nonlinear expansion layer, an attention mechanism, and a MIMO mechanism; (3) dedicated neural network design for adaptive identification of traffic flow model parameters. A traffic state estimator built on the basis of this extended PI-DeepONet architecture was evaluated with respect to a short freeway stretch of NGSIM and a large-scale urban expressway in China, along with other four baseline TSE methods. The evaluation results demonstrated that this novel TSE method outperformed the baseline methods with high-precision estimation results of flow and mean speed.

*Hongxin Yu, Yibing Wang, Fengyue Jin, Meng Zhang, and Anni Chen*

**INTRODUCTION**

Traffic state estimation (TSE) is crucial for advanced road traffic surveillance and control. TSE aims to utilize a limited amount of traffic sensing data to estimate traffic variables such as speed and flow in an adequate spatial resolution within a road network in real-time.

Freeway TSE studies can be categorized as model-driven and data-driven based on their methodological foundations. Model-driven TSE relies on macroscopic traffic flow model, e.g. the first-order LWR *(1)* or higher-order model PW *(2)*, originated from hydrodynamics, to describe spatiotemporal dynamic evolution of traffic flow. The related works applies various filtering techniques *(3, 4)* to fuse real-time sensing data with traffic flow models to deliver network traffic state estimates. Macroscopic traffic flow models can be extensively calibrated with real-world data for fidelity. Therefore, model-driven TSE results provide good explainability, though the estimation accuracy is limited due to some inherent errors and limitations of the models.

On the other hand, data-driven TSE leverages historical sensing data and employs statistical models *(5, 6)* or machine learning methods *(7, 8)* to unveil spatiotemporal correlations for the inference of traffic state distribution and evolution. Purely data-driven TSE methods do not rely on analytical mathematical models, and normally lack transparency due to their "black-box" nature, in addition to limited accuracy.

In order to take advantage of both model-driven and data-driven methods of TSE, increasingly more attentions have been paid to data-model dual-driven approaches, especially those based on physics-informed neural network (PINN) *(9)*. PINN first emerged in the field of scientific computing, particularly for solving partial differential equations (PDEs) *(10)*. The main idea of PINN is to impose physical models as constraints on the employed deep-learning neural networks and guide the neural computing to follow physical laws so as to improve the neural networks' understanding of complex system dynamics. Macroscopic traffic flow models are a specific type of PDEs, and PINN has recently been exploited for freeway TSE *(11, 12)*. However, due to limitations in the network architecture of PINN, the generalization performance of PINN-based TSE is rather weak: once the traffic environment changes, re-training is inevitable. This work seeks for help from a more robust method of physics-informed deep operator network (PI-DeepOnet) *(13)*.

In mathematics, a function maps from its domain to range, while an operator is a mapping from a function space to another function space. An operator can be of the explicit type such as the integral, Laplace transform, or of the inexplicit type like solution operators of PDEs. TSE can actually be viewed as an operator that maps available measurements of traffic state variables into unmeasured traffic state variables in real time. Just as various approaches of deep neural networks (DNNs) are employed for function regression *(10)*, DNNs can also be based for operator approximation, e.g. fourier neural operator *(14)* and graph neural operator *(15)*. Lu et al. *(16)* proved rigorously that DNNs are universal approximators for nonlinear continuous operator, based on which Lu et al. *(16)* designed a deep operator network (DeepONet) and demonstrated that it can learn various explicit and inexplicit operators quite accurately. As inspired by the idea of DeepONet and the potential of PINN, Wang et al. *(13)* proposed PI-DeepONet and applied it to deal with wave propagation, reaction-diffusion dynamics, and stiff chemical kinetics. For the first time this paper has designed a freeway traffic state estimator in the idea of PI-DeepONet.

The architecture of PI-DeepONet consists of four parts, branch network, trunk network, model-based computational graph, and loss function. The branch network of the DNN type is used to encode the initial condition space, which corresponds to real-time traffic measurements in terms of freeway TSE, and the trunk network of DNN is used to encode the domain of the output functions, e.g. the locations of interest for TSE. The model-based computational graph imbeds physical knowledge into the architecture to determine the physical loss in order for PI-DeepONet's output to be consistent with the employed physical model. In addition, the discrepancy between real observations and PI- DeepONet's outputs is evaluated to deliver the supervised loss. The sum of physical loss and supervised loss is based to optimize PI-DeepONet's weight parameters.

Albeit with considerable potential, the original PI-DeepONet architecture exhibits the following limitations that hinder the direct application of PI-DeepONet to TSE:
- it is with so simple a network structure that the branch and trunk networks accept 1-D data input only, and thus cannot handle spatiotemporal traffic data that typically forms a manifold on a 2-D (space-time) domain;
- it operates in a single-input single-output manner, making it hard to handle TSE for multiple traffic flow variables simultaneously (e.g. flow and speed);
- it does not allow the parameters of the imbedded physical model to change over time, an indispensable property for the environmentally adaptive TSE.

In order to address these limitations for the sake of TSE, we have proposed an extended PI-DeepONet architecture:
- The fully connected branch neural network is replaced with a convolutional neural network (CNN) to accommodate 2-D data input required, enabling the extraction of spatiotemporal features from historical traffic data. Additionally, a nonlinear expansion layer and an attention mechanism are introduced to the trunk network to enhance its ability in handing complex nonlinear systems.
- A multi-input multi-output (MIMO) mechanism is proposed for the estimation of multiple traffic flow variables.
- A parameter estimation neural network is designed for adaptive identification and evolution tracking of traffic flow model parameters.

A comprehensive performance evaluation has been conducted with respect to a short freeway stretch based on the Next Generation Simulation (NGSIM) data set and a long urban expressway. The evaluation has demonstrated the designed traffic state estimator is superior to all baseline TSE methods considered, and in particular, the extension and enhancement made with regard to the original PI-DeepONet significantly improves estimation accuracy and generalization capability of the traffic state estimator. To the best of our knowledge, it is the first time to apply either the original PI-DeepONet or its extended architecture to TSE tasks.

**Original PI-DeepONet**
**Preliminaries**

Let $L$ and $T$ represent the spatial and temporal lengths of a targeted space-time domain $[0, L] \times [t_0, t_0 + T]$. Consider a point $(x, t)$ in the domain. Let $s(x, t)$ represent the spatiotemporal dynamics of an interested physical state at $(x, t)$. Consider that $s(x, t)$ can be described by the following nonlinear partial differential equations (PDEs) system:

$$\mathcal{N}(s(x,t); \lambda) = 0, \quad x \in [0, L], \quad t \in [t_0, t_0 + T], \quad (1)$$
$$s(x, t_0) = u(x, t_0), \quad x \in [0, L], \quad (2)$$

where $\mathcal{N}$ is the nonlinear partial differential operator, $\lambda$ represents system parameters, $u$ represents the initial condition. In other words, $\mathcal{N}$ maps from $u(x, t_0)$ to $s(x, t)$, $t \in [t_0, t_0 + T]$. Freeway traffic flow is one case of the system described by **Equations 1** and **2**, whereby $u(x, t_0)$ corresponds to traffic measurements from sensor locations $x_1, x_2, \ldots$ at each current time instant $t_0$. TSE aims to determine the whole state $s(x, t_0)$ based on a limited amount of traffic measurements $u(x_1, t_0)$, $u(x_2, t_0), \ldots$ for any $t_0$.

Consider two infinite-dimensional Banach spaces: the initial condition space $\mathcal{U}$ and the solution mapping space $\mathcal{S}$. PI-DeepONet aims to approximate the operator $\mathcal{N}$. To incorporate $\mathcal{N}(s(x,t); \lambda)$ into the neural network computing, we introduce the physical residual $f(x, t; \lambda)$ as follows:

$$f(x, t; \lambda) = \mathcal{N}(\hat{s}(x, t); \lambda), \quad (3)$$

where $\hat{s}(x,t)$ denotes the approximation of $s(x,t)$ resulting from the neural network computing, and if $\hat{s}(x,t) = s(x,t)$, $f(x,t;\lambda)$ is zero; otherwise, it is not and hence called the physical residual *(10)*.

**PI-DeepONet Architecture**
As displayed in **Figure 1**, the architecture of original PI-DeepONet *(13)* parameterized by $\theta$ consists of four parts, branch network, trunk network, model-based computational graph, and loss function. The branch network takes as input a set of initial conditions $\boldsymbol{u} = [u(x_1,t_0), u(x_2,t_0), \cdots, u(x_W,t_0)]$, which means in terms of TSE real-time traffic measurements from sensors $\{x_i\}_{i=1}^W$, with $W$ being the number of sensors and $x_i$ the $i$-th sensing location.

The branch network returns a feature $[b_1(\boldsymbol{u}), b_2(\boldsymbol{u}), \cdots, b_K(\boldsymbol{u})]$, where $K$ is the number of encoded features. Given the set of initial conditions at $t_0$, the spatiotemporal coordinate point $(x, t_0 + t)$, $0 \leq t \leq T$, is written in this paper as $\boldsymbol{y} = (x,t)$, which is also called the query coordinate. The trunk network takes any query coordinate as input and returns feature $[t_1(\boldsymbol{y}), t_2(\boldsymbol{y}), \cdots, t_K(\boldsymbol{y})]$. Accordingly, $u(x,t_0)$ can be written briefly as $u(x)$. Then, the estimated value at $(x,t)$, given the initial condition $\boldsymbol{u}(x)$, is calculated as the Hadamard product of the branch and trunk network features:

$$G_\theta(\boldsymbol{u})(\boldsymbol{y}) = \sum_{k=1}^{K} b_k(\boldsymbol{u}) t_k(\boldsymbol{y}) = \sum_{k=1}^{K} b_k\big(u(x_1), u(x_2), \cdots, u(x_w)\big) t_k(x,t), \tag{4}$$

where $\theta$ represents all trainable weights and biases in the branch and trunk neural networks. $G_\theta(\boldsymbol{u})(\boldsymbol{y})$ is the estimated state value at the query coordinate $\boldsymbol{y} = (x,t)$ given the set of initial conditions $\boldsymbol{u}(x)$.

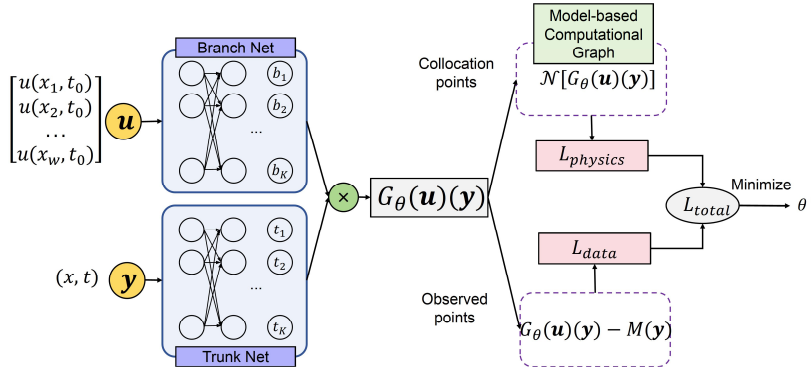

**Figure 1 Structure of original PI-DeepONet**

For the purpose of PI-DeepONet traning, two loss functions are constructed. Firstly, the model-based computational graph determines the physical residual in order to construct the physical loss $\mathcal{L}_{physics}$ in **Equation 5**, aligning the PI-DeepONet's output with the employed physical model.

$$\mathcal{L}_{physics}(\theta) = \frac{1}{N \times P} \sum_{i=1}^{N} \sum_{j=1}^{P} \big|\mathcal{N}\big(G_\theta(\boldsymbol{u}^i)(\boldsymbol{y}_c^{i,j}); \lambda\big)\big|^2 \tag{5}$$

where $i$ denotes the index of the $i$-th current/initial time instant, $\boldsymbol{u}^i = [u(x_1,t_0^i), u(x_2,t_0^i), \cdots, u(x_W,t_0^i)]$, and $N$ is the total number of initial time instants. Note that, in terms of TSE, $\boldsymbol{u}^i$ corresponds to sensing data, so $N$ actually refers to the size of historical sensing data set used for the training of PI-DeepONet.

For each initial time instant $t_0^i$, consider a space-time domain $\mathcal{D} = [0, L] \times [t_0^i, t_0^i + T]$, and take a number $P$ of collocation points from $\mathcal{D}$ randomly, with each collocation point in the form of $(x, t_0^i + t)$, $x \in [0, L]$, $t \in [0, T]$. Recalling the definition of the query coordinate $y$, denote by $y_c^{i,j}$ the $j$-th collocation point given $u^i$. Then, all collocation points are used to calculate the physical residuals $\mathcal{N}(G_\theta(u^i)(y_c^{i,j}); \lambda)$ based on **Equation 3**, and furthermore the physical loss $\mathcal{L}_{physics}$ via **Equation 5**. The partial derivatives involved in $\mathcal{N}$ with respect to $y$ can be computed using automatic differentiation for neural network.

As stated above, all collocation points are randomly taken from the space-time domain $[0, L] \times [t_0^i, t_0^i + T]$, without considering whether a collocation point refers to a sensor location or not. Next, we focus on the sensor locations and check the consistency or discrepancy between the estimates and sensor observations at those locations.

Let $R$ denote the number of query coordinates $y$ with observed true values (named observed points, defined as $y_o$) within the domain $[0, L] \times [t_0^i, t_0^i + T]$ given initial conditions $u^i$. Then, the supervised loss is constructed as follows:

$$\mathcal{L}_{data}(\theta) = \frac{1}{N \times R} \sum_{i=1}^{N} \sum_{r=1}^{R} |G_\theta(u^i)(y_o^{i,r}) - M(y_o^{i,r})|^2 \tag{6}$$

where $y_o^{i,r}$ represents the $r$-th observed point within $[0, L] \times [t_0^i, t_0^i + T]$, $M(y_o^{i,r})$ the true observations at $y_o^{i,r}$, and $R$ the total number of observed points. Assuming the sensing time interval is exactly $T$ in the case of TSE, $R$ is twice the number of sensors.

Overall, the total loss function reads:

$$\mathcal{L}_{total}(\theta) = \alpha_1 \mathcal{L}_{data}(\theta) + \alpha_2 \mathcal{L}_{physics}, \tag{7}$$

where $\alpha_1$ and $\alpha_2$ are the weights for the respective loss terms.

Provide all initial conditions $u^i$ along with their corresponding values $y_c^{i,j}$, $y_o^{i,r}$, and $M(y_o^{i,r})$ to form input-output data pairs and generate a training dataset. We can utilize neural network optimizers such as Adam to obtain the optimal neural network parameters $\theta^* = \underset{\theta}{\text{argmin}}\, \mathcal{L}_{total}(\theta)$.

**Extended PI-DeepONet**
**General description**
This setion develops in the idea of PI-DeepONet a physics-guided deep learning architecture for the design of a freeway traffic state estimator in **Figure 2**.

As shown in **Figure 1**, the original PI-DeepONet employs a basic fully connected network (FCN) that only accepts 1-D (vector-like) input and operates in a single-input single-output manner (i.e., both input $u$ and output $G_\theta(u)(y)$ are scalar, each addressing one specific variable). Unfortunately, this is quite inconvenient for performing TSE.

Firstly, the use of 1-D data input neglects the temporal characteristics of data, and utilizing 2-D data input as displayed in **Figure 2**, with one dimension addressing traffic flow variables and the other the timeline, facilitates the applicaiton of more matured approaches to traffic state estimation and prediction, like convolutional neural networks (CNNs), and can lead to quite accurate traffic state estimation and prediction results *(17)*.

Secondly, traffic flow condition at a spot on the road is much better described by two of the three fundamental variables flow, mean speed, and density rather than one variable. Also, spot traffic sensors on roads often deliver both flow and mean speed measurements together. As such, the PI-DeepONet-based traffic state estimator is expected to accept a multi-variable input (traffic measurements) and deliver a multi-variable output (traffic state estimates).

In addition, the original PI-DeepONet assumes that the physical model parameters are fixed at some apriori parameters, overlooking impacts of the spatiotemporal imhomogneity of traffic flow if applying to TSE. To make up for this deficiency, the extended architecture should allow the parameters of the imbedded physical model to change over time so as to enable the adaptation of designed traffic state estimator to the environmental changes.

With the above in mind, this section proposes an extended architecture of PI-DeepONet as shown in **Figure 2**, which has the following innovative features: (1) CNN-based branch network; (2) trunk network enhanced with an attention mechanism and a nonlinear expansion layer; (3) multi-input multi-output (MIMO) enabling module; (4) parameter network that self-learns traffic flow model parameters.

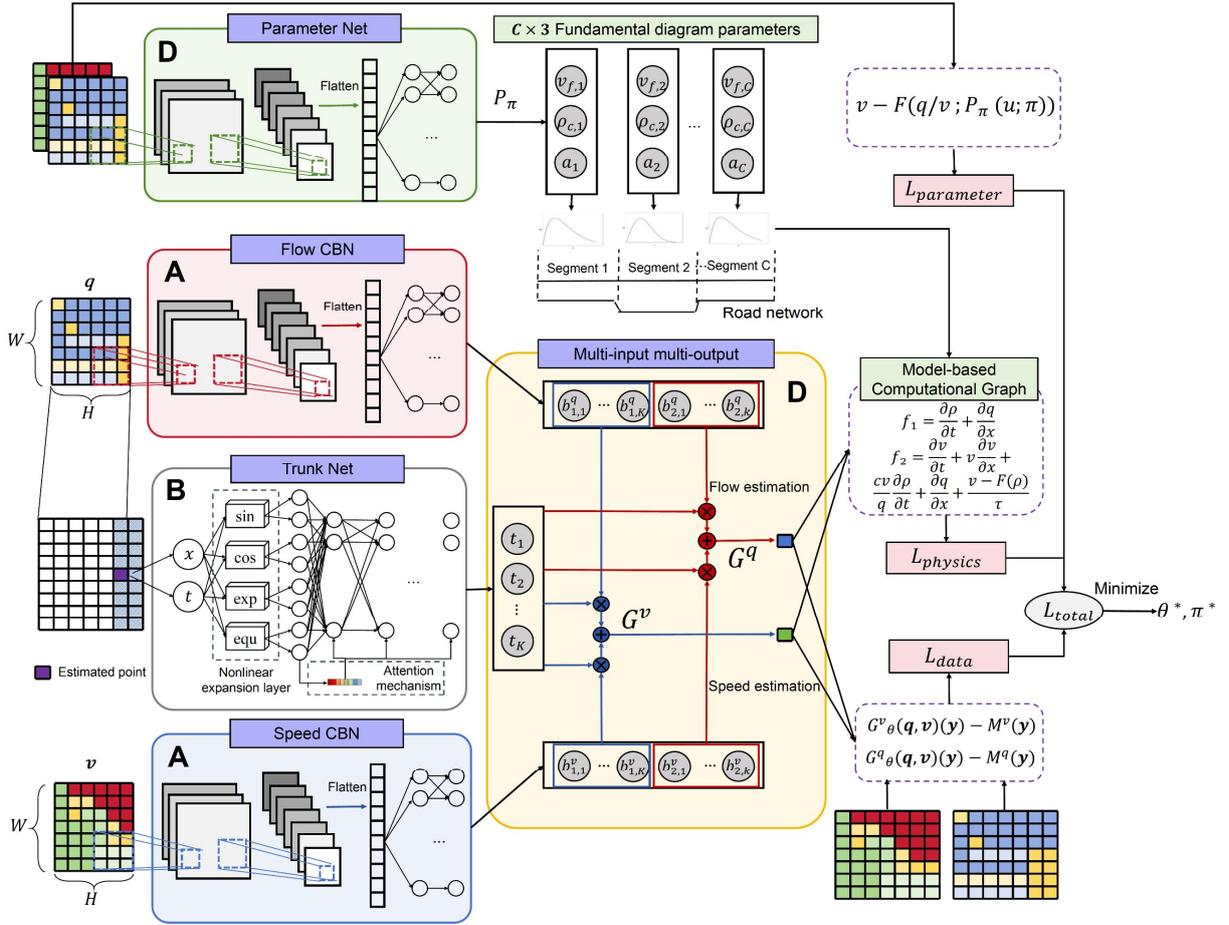

**Figure 2** Structure of extended PI-DeepONet: (A) CNN-based branch network; (B) Trunk network enhanced with attention mechanism and nonlinear expansion layer; (C) Multi-input multi-output mechanism; (D) Parameter network

**CNN-based branch network (CBN)**
Consider a number $W$ of sensors installed in a freeway network. Traffic sensing data set often includes not only measurements at each current time instant, but also (historical) measurements from the previous number $H-1$ of time instants. Therefore, the input data can be arranged as a matrix of the dimension $(H, W)$. Following the convention in the field of machine learning, the vector-like input in **Figure 1** is referred to as 1-D, while the matrix-like input in **Figure 2** is called 2-D, which is very convenient for running CNNs in the brank network for the sake of traffic state esitmaiton and prediction.

As shown with block A in **Figure 2**, CBN comprises several convolution layers *(18)* that extract spatiotemporal features from the input data, and a flatten layer is followed to convert the data into 1-D format. Then, an FCN is applied to yield an output encoding with dimension of $2 \times K$. The CBN process can be expressed as:

$$\text{CBN}: \mathbb{R}^{H \times W} \to \mathbb{R}^{2 \times K} \tag{8}$$

**Trunk network enhanced with attention mechanism and nonlinear expansion layer**
As shown with block B in **Figure 2**, the original trunk network is extended by introducing a nonlinear expansion layer and an attention mechanism.

The nonlinear expansion layer aims to enhance its input by use of nonlinear functions, thereby improving the trunk network's nonlinear approximation capability. This nonlinear extension layer performs an affine transformation on the original two-dimensional input, projecting it into an eight-dimensional one:

$$Z^{(0)}[\mathbf{y}] = [cos(x,t), sin(x,t), exp(x,t), (x,t)], \tag{9}$$

where $cos(x,t)$, $sin(x,t)$, and $exp(x,t)$ denote cosine, sine, and exponential functions, respectively, and $Z^{(0)}$ is the output of the nonlinear extension layer.

Additionally, an improved FCN with an embedded attention mechanism is introduced into the trunk network to enable neural networks to "focus on key points" by learning and grasping essential information, thereby enhancing model performance and generalization capabilities. The attention mechanism has been proven to offer better performance than standard FCNs *(19)*. The forward propagation process of the mechanism is as follows:

$$U = \phi(Z^{(0)} w^u + b^u), \quad V = \phi(Z^{(0)} w^v + b^v) \tag{10}$$
$$H^{(1)} = \phi(Z^{(0)} w^{(1)} + b^{(1)}) \tag{11}$$
$$Z^{(l)} = \phi(H^{(l)} w^{(l)} + b^{(l)}), \quad l = 1, \dots, L \tag{12}$$
$$H^{(l+1)} = (1 - Z^{(l)}) \odot U + Z^{(l)} \odot V, \quad l = 1, \dots, L \tag{13}$$

In **Equation 10**, $\odot$ denotes the Hadamard product, $\phi$ denotes the Tanh activation function, $w^u$, $w^v$ and $b^u$, $b^v$ are the weights and biases for $Z^{(0)}$, respectively. $U$ and $V$ are the attention weights determined by the $Z^{(0)}$, i.e. extracting attention weights from the nonlinear expansion layer. $w^{(l)}$ and $b^{(l)}$ are the weights and biases for each layer of the *l*-th hidden layer, respectively; $L$ the number of hidden layers. **Equation 11** provides the input for the first hidden layer. In **Equation 12**, $Z^{(l)}$ is the output of the *l*-th hidden layer. In **Equation 13**, $H^{(l+1)}$ is the input for the next hidden layer $l + 1$, which applies the attention weights obtained from Equation 10 to $Z^{(l)}$, producing a weighted input.

**Multi-input multi-output mechanism**
As inspired by the literature *(20)*, we adopt a MIMO mechanism to achieve multiple state outputs for the sake of TSE. The network structure is shown with block C in **Figure 2**.

Firstly, two CBN networks, Speed CBN and Flow CBN, are constructed. Their inputs are speed features $\mathbf{v}$ and traffic flow features $\mathbf{q}$ with dimension of $H \times W$, respectively, and their outputs are $2 \times K$ dimensional speed encoded features $b(\mathbf{v}) = [b^v_{1,1}(\mathbf{v}), \cdots, b^v_{1,K}(\mathbf{v}), b^v_{2,1}(\mathbf{v}), \cdots, b^v_{2,K}(\mathbf{v})]$ and flow encoded features $b(\mathbf{q}) = [b^q_{1,1}(\mathbf{q}), \cdots, b^q_{1,K}(\mathbf{q}), b^q_{2,1}(\mathbf{q}), \cdots, b^q_{2,K}(\mathbf{q})]$, respectively. Integrated with the trunk network output $[t_1(\mathbf{y}), t_2(\mathbf{y}), \cdots, t_q(\mathbf{y})]$, the final output of the operator network reads:

$$\begin{cases} G_\theta^v(\boldsymbol{q},\boldsymbol{v})(\boldsymbol{y}) = \sum_{k=1}^{K} b_{1,k}^v(\boldsymbol{v})t_k(\boldsymbol{y}) + \sum_{k=1}^{K} b_{1,k}^q(\boldsymbol{q})t_k(\boldsymbol{y}) \\ G_\theta^q(\boldsymbol{q},\boldsymbol{v})(\boldsymbol{y}) = \sum_{k=1}^{K} b_{2,k}^v(\boldsymbol{v})t_k(\boldsymbol{y}) + \sum_{k=1}^{K} b_{2,k}^q(\boldsymbol{q})t_k(\boldsymbol{y}) \end{cases}, \quad (14)$$

where $G_\theta^v$ and $G_\theta^q$ are the estimated values for mean speed and flow, respectively.

**Parameter network for self-learning traffic flow model parameters**
Previous research *(3, 21, 22)* has demonstrated that (1) traffic flow exhibits spatiotemporal inhomogeneity, i.e. traffic flow model parameters (such as free-flow speed $v_f$, critical density $\rho_c$, capacity $q_c$, etc) may quite likely change in time (typically during the day vs. at night) and in space (typically due to the presence of slope, curvure, tunnel, bridge, lane drop, etc); (2) traffic flow inhomogeneity has a significant impact on the performance of traffic flow modeling and TSE; (3) online model parameter estimation is required in order to ensure satisfactory TSE results.

In the light of this, we have designed a dedicated network for adaptive model parameter identification based on traffic masurements. As shown with block D in **Figure 2**, this parameter network parameterized by $\pi$ adopts the same network structure and input as CBN with the final output dimension being $C \times 3$, corresponding to the traffic fundamental diagram parameters for each road segment. To this end, a corresponding loss function $\mathcal{L}_{parameter}$ is introduced as follows:

$$\mathcal{L}_{parameter}(\pi) = \frac{1}{N \times Q} \sum_{i=1}^{N} \sum_{r=1}^{R} \left| v^{i,r} - F\left(\frac{q^{i,r}}{v^{i,r}}; P_\pi(\boldsymbol{u}^i; \pi)\right) \right|^2, \quad (16)$$

where $P_\pi$ is the estimated parameter value, $v^{i,r}$ and $q^{i,r}$ represent the speed and flow corresponding to the $r$-th observation point under the $i$-th initial condition set $\boldsymbol{u}^i$. $\mathcal{L}_{parameter}$ evaluates the accuracy of the fundamental-diagram-based speed estimates. $\mathcal{L}_{parameter}$ is used to adjust the weights of the parameter network, and also incorporated into the total loss function.

**Model computational graph**
For the model computational graph, this study applies the higher-order traffic flow model PW *(2)*, which is specifically formulated as follows:

$$\begin{cases} \dfrac{\partial \left(\dfrac{q}{v}\right)}{\partial t} + \dfrac{\partial q}{\partial x} = 0 \\ \dfrac{\partial v}{\partial t} + v\dfrac{\partial v}{\partial x} + \dfrac{cv}{q}\dfrac{\partial \left(\dfrac{q}{v}\right)}{\partial x} + \dfrac{v - F\left(\dfrac{q}{v}\right)}{\tau} = 0 \end{cases} \quad (17)$$

In **Equation 17**, $\tau$ is the relaxation time parameter, and $c$ characterizes driver expectations. In this study, the two parameters were fixed at 18 and 40. The two corresponding physical residuals are as follows:

*Hongxin Yu, Yibing Wang, Fengyue Jin, Meng Zhang, and Anni Chen*

$$\begin{cases} f_1 = \frac{\partial(\frac{q}{v})}{\partial t} + \frac{\partial q}{\partial x} \\ f_2 = \frac{\partial v}{\partial t} + v\frac{\partial v}{\partial x} + \frac{cv}{q}\frac{\partial(\frac{q}{v})}{\partial x} + \frac{v - F(\frac{q}{v})}{\tau} \end{cases} \quad (18)$$

which is applied to construct the physical loss for the network training.

**SIMULATION INVESTIGATIONS**

This section evaluates the performance and generalization capability of the designed traffic state estimator based on the extended PI-DeepONet architecture. To this end, two datasets were considered: (1) trajectory data of NGSIM collected from a short freeway stretch in USA; (2) fixed-sensing data from a large-scale urban expressway in China. In addition to the TSE method proposed in this paper, four baseline methods were considered for comparison:

- **Linear Interpolation (Inter2d)** delivering TSE results based on linear interpolation with adjacent data points *(5)*.
- **Adaptive Smoothing (AS)** considering traffic wave speeds under free-flow and congested conditions reproducing spatiotemporal traffic state using a smoothing kernel filter to *(6)*.
- **Physics-Informed Neural Network (PINN)** based on the PW model *(12)*.
- **PI-DeepONet with** the MIMO mechanism based on the PW model.

To evaluate the performance of the TSE methods, two metrics were selected: root mean square error (RMSE) and relative error (RE). Their calculation formulae are as follows:

$$\text{RMSE} = \sqrt{\frac{1}{n}\sum_{i=1}^{n}(\hat{y}_i - y_i)^2} \quad (19)$$

$$\text{RE} = \frac{\sqrt{\sum_{i=1}^{n}(\hat{y}_i - y_i)^2}}{\sqrt{\sum_{i=1}^{n}y_i^2}} \quad (20)$$

**A Short Freeway Stretch**

The considered short freeway stretch of 1600 feet is part of I-80, Emeryville, California. Vehicle trajectory data of 2700 seconds collected on the stretch constitutes part of the NGSIM dataset *(23)*. To facilitate the performance evaluation and result analysis, the trajectory data was aggregated into traffic flow and mean speed data, with a temporal resolution of 5 seconds and a spatial resolution of 20 feet.

The dataset was divided into three parts in the data quantity ratio of 7:1:2, and used for training, validation, and testing of the traffic state estimator, respectively. To conduct the TSE task, the sensing data was available every 160 feet, and used as input for the training process. Data from the locations every 20 feet that were not involved in network training were used for the performance evaluation.

*Overall Result*

The evaluation performance is presented in **TABLE 1**. As shown, the TSE method developed in the extended PI-DeepONet architecture outperforms the other four (baseline) methods. For flow estimation, the RMSE index is 0.354 veh/s, and the RE index is 12.57%. For speed estimation, RMSE is 3.64 ft/s, and RE is 9.51%. The estimation results of extended PI-DeepONet are superior not only to traditional methods of Inter2d and AS, but also to the model-data dual-driven methods PINN and PI-DeepONet.

In association with **TABLE 1**, some detailed estimation results of extended PI-DeepONet are presented in **Figures 3** and **4**. **Figure 3** plots a heatmap of the global TSE results versus the ground truth, while **Figure 4** showcases the TSE results in the spatial resolution of 200 feet. Clearly, the TSE results satisfactorily reproduce the process of congestion propagation and dissipation along the stretch, demonstrating that the traffic state estimator has gained the capability of tracking the spatiotemporal evolution of traffic flow dynamics by learning from both historical data used and traffic flow model embedded.

*Sensor Sensitivity Analysis*
The results in **TABLE 1** as well as **Figures 3** and **4** were based on a configuration of 11 input sensors. **Figure 5** examines the impact of the number of input sensors on the TSE performance. As shown, among the compared methods, the extended PI-DeepONet exhibited the lowest sensitivity in flow and speed estimates. It employs a special data-model-fusion architecture. When the number of input sensors is small, it relies more on the model to infer traffic state; when the number of sensors is ample, it uses the data to optimize the model.

**TABLE 1 TSE Performance of Different Methods on the Short Freeway Stretch Dataset**

| Method | Performance | | | |
| --- | --- | --- | --- | --- |
| | RE of speed (%) | RMSE of speed (ft/s) | RE of flow (%) | RMSE of flow (veh/s) |
| Inter2d | 12.68 | 4.85 | 12.80 | 0.360 |
| AS | 11.62 | 4.45 | 14.91 | 0.419 |
| PINN | 10.25 | 3.92 | 12.95 | 0.364 |
| PI-DeepONet | 11.07 | 4.23 | 13.35 | 0.375 |
| Extended PI-DeepONet | **9.51** | **3.64** | **12.57** | **0.354** |

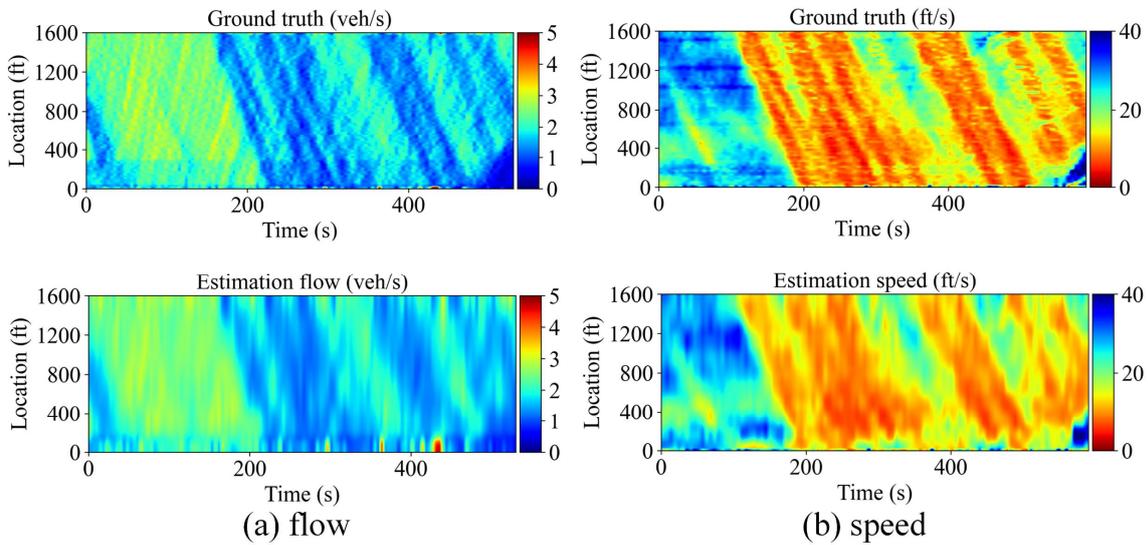

(a) flow      (b) speed

**Figure 3 Heatmaps for TSE results of extended PI-DeepONet for the short freeway stretch: (a) flow; (b) speed**

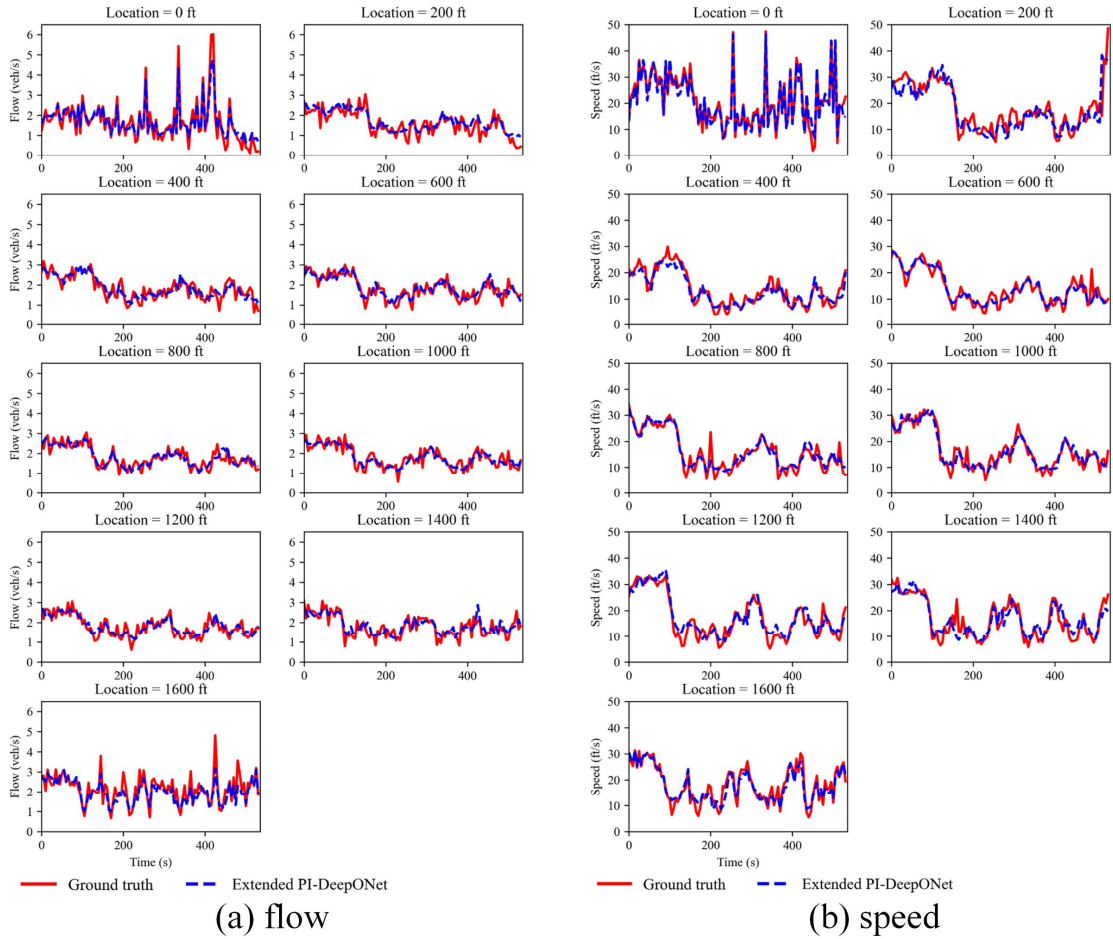

(a) flow  (b) speed

**Figure 4 TSE results of extended PI-DeepONet for the short freeway stretch in the spatial resolution of 200 feet s: (a) flow; (b) speed**

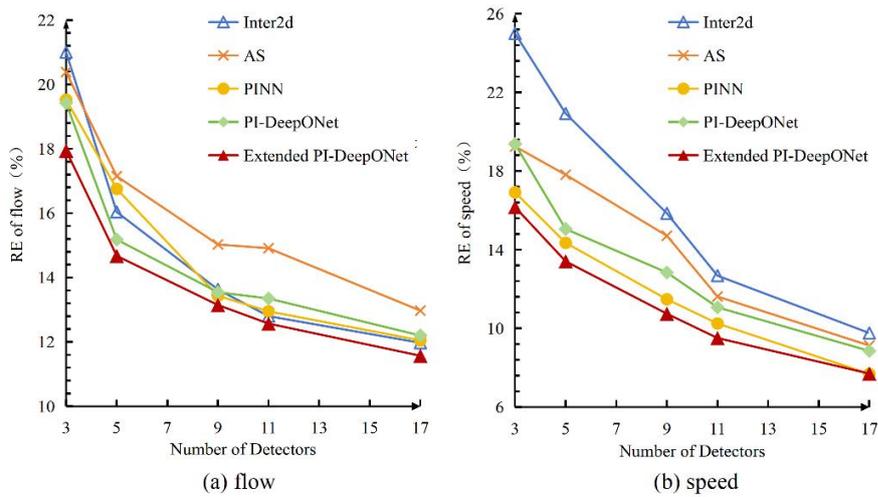

(a) flow  (b) speed

**Figure 5 Sensitivity of TSE performance to the number of input sensors: (a) flow; (b) speed**

Hongxin Yu, Yibing Wang, Fengyue Jin, Meng Zhang, and Anni Chen

**A Large-scale Urban Expressway**
Given its short 490-meter length, the freeway segment's NGSIM data exhibited limited traffic flow complexity, posing limited challenge to the designed traffic state estimator. In order to thoroughly evaluate the performance of the proposed new TSE method along with the four baseline methods, a large-scale urban expressway example was also considered.

**Figure 6** depicts the expressway, which extends about 17.47 km and is equipped with 32 sensors (D1-D32) for flow and mean speed data in the time resolution of 5 minutes. The utilized data was collected from June 1 to June 30, 2019. In the light of dynamic system observability *(24)*, 22 mainline sensors marked in red in **Figure 6** were chosen as the input sensors for TSE, while the remaining 10 sensors in black served as the evaluation sensors. The dataset was divided into training, validation, and testing parts in the quantity ratio of 7:1:2.

The evaluation performance is given in **TABLE 2**. Extended PI-DeepONet still achieved the best performance.

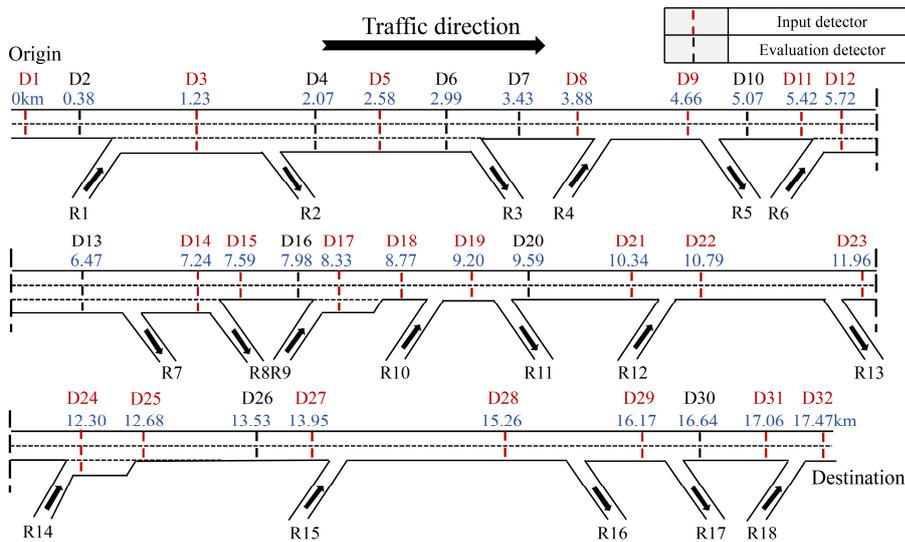

**Figure 6 Topology of the experimental large-scale road network**

**TABLE 2 TSE Performance of Different Methods on the Large-scale Road Network Dataset**

| Method | Performance | | | |
| --- | --- | --- | --- | --- |
| | RE of speed (%) | RMSE of speed (km/h) | RE of flow (%) | RMSE of flow (veh/h) |
| Inter2d | 12.01 | 11.18 | 13.46 | 332.73 |
| AS | 12.12 | 11.30 | 12.43 | 307.26 |
| PINN | 10.17 | 9.48 | 10.27 | 254.02 |
| PI-DeepONet | 10.60 | 9.91 | 12.56 | 310.67 |
| Extended PI-DeepONet | **8.99** | **8.38** | **7.76** | **191.80** |

Like the short freeway stretch case, the extended PI-DeepONet methods surpasses all other baseline methods in the TSE performance. For flow estimation, the traditional statistical methods of Inter2d and AS struggle to capture flow changes caused by the freeway topology depicted in **Figure 6**, leading to significant errors in the mainline flow estimates, which eventually affected the mean speed estimates as well. For speed estimation, neither of the two methods Inter2d and AS takes advantage of established knowledge on traffic flow dynamics, especially traffic kinematic wave and shockwave, and hence cannot track complex traffic flow dynamics well.

On the other hand, either of PINN and PI-DeepONet includes traffic flow models in their architecture, which have the targeted freeway topology recorded and corresponding traffic flow dynamics formulated, and thus can deliver better estimates of flow and mean speed. On the top of this, even better TSE results were obtained under the extend PI-DeepONet architecture, demonstrating that the innovative features introduced via the extension from the original PI-DeepONet, i.e. the acceptance of 2-D data input so as to support CNN-based computations, the MIMO and attention mechanisms, adaptive identification of traffic flow model parameters, play a significant role.

More detailed estimation results of extended PI-DeepONet are presented in **Figures 7** and **8**. **Figure 7** plots a heatmap of the global TSE results versus the ground truth, while **Figure 8** presents the TSE results at all evaluation sensors. It can be seen that extended PI-DeepONet effectively captures the congestion propagation in several congested areas, particularly the sharp speed drop and stop-and-go waves at evaluation sensors 3-9.

Finally, to check the difference between PI-DeepONet and extended PI-DeepONet, **Figure 9** compares the frequency histograms of relative errors at all evaluation sensors over the study time horizon, where the brown part is shared by both cases. As displayed, clearly extended PI-DeepONet exhibited a notably higher frequency in the high-accuracy (light blue) region, particularly when the relative errors were below 5%.

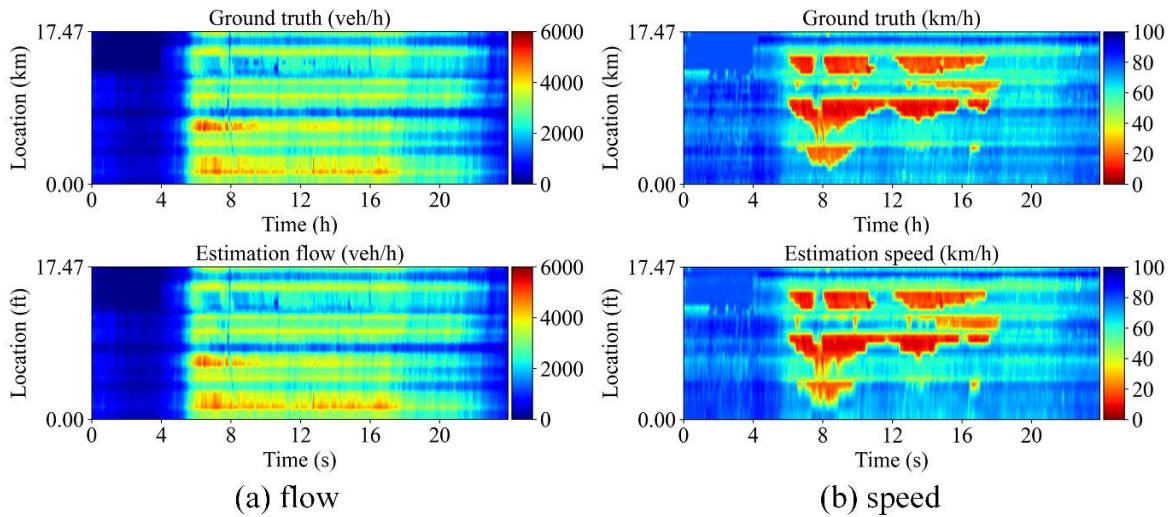

**Figure 7 Heatmaps for TSE results of extended PI-DeepONet for the large-scale urban expressway: (a) flow; (b) speed**

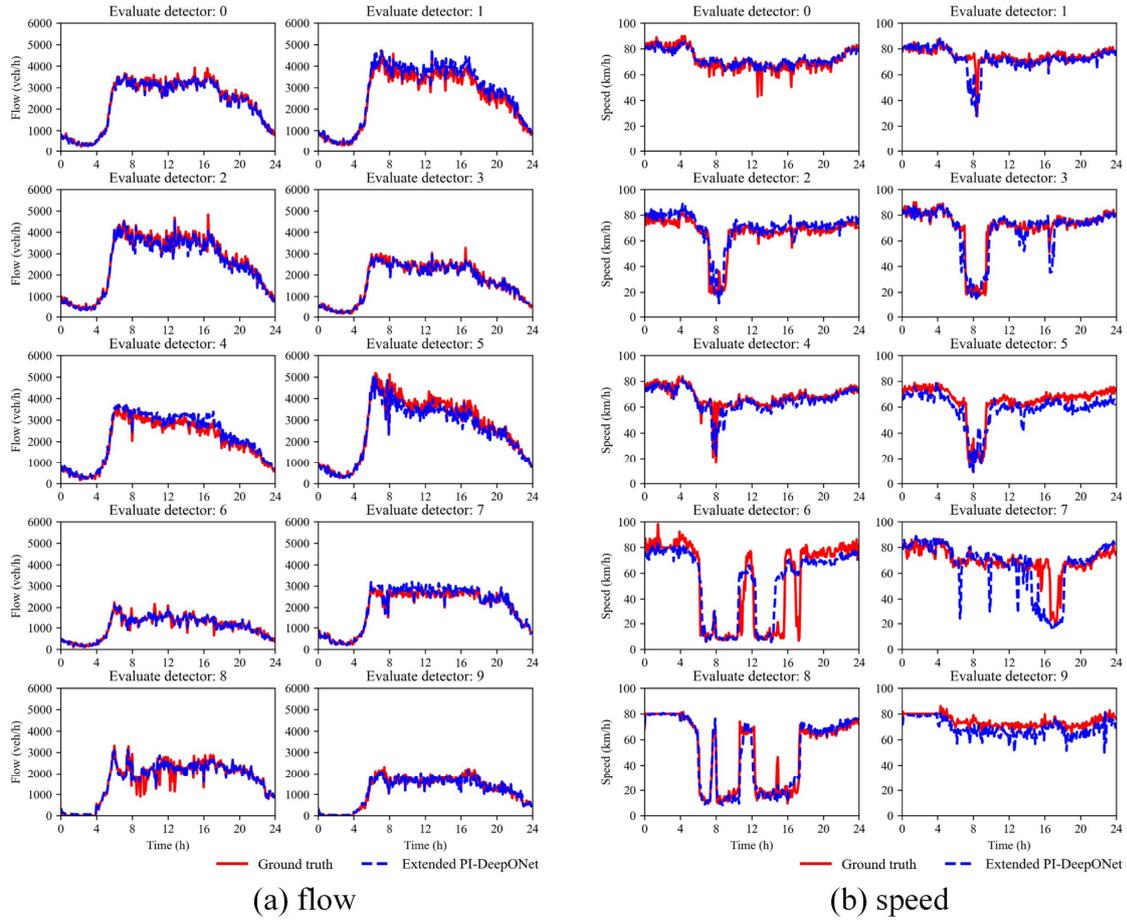

(a) flow  (b) speed

**Figure 8** TSE results of extended PI-DeepONet for the large-scale urban expressway: (a) flow; (b) speed

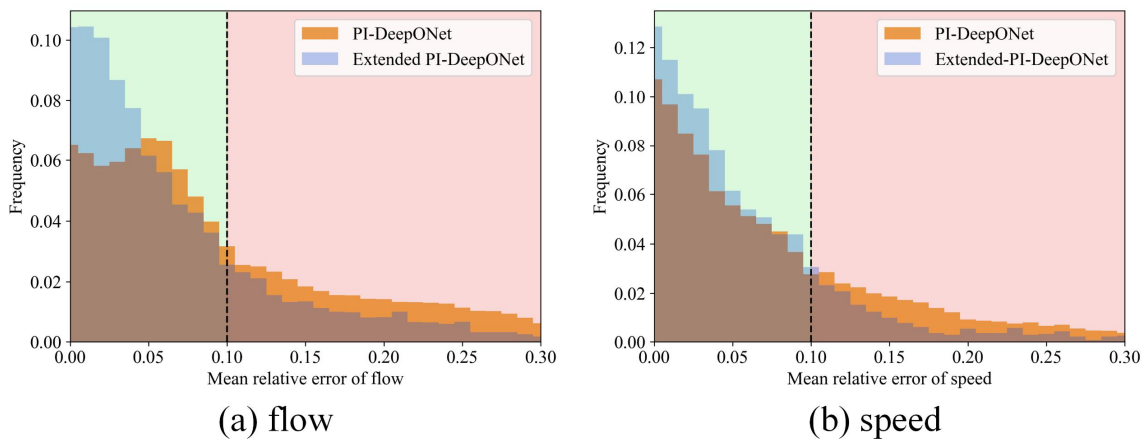

(a) flow  (b) speed

**Figure 9** Frequency histograms of relative errors at all evaluation sensors over the study time horizon: (a) Flow; (b) Speed

**CONCLUSIONS**

      For the first time this paper has proposed to study real-time freeway traffic state estimation in the idea of PI-DeepONet and developed an extended architecture based on the original PI-DeepONet for the sake of freeway traffic state estimation. The extended PI-DeepONet method has been evalauted on real data from a short freeway stretch and a large-scale urban expressway, delivering satisfactory estimation results in both cases. As a followup work, the study will be further extended to address freeway traffic state prediction over a future time horizon.

**AUTHOR CONTRIBUTIONS**
The authors confirm contribution to the paper as follows: study conception and design: Hongxin Yu, Yibing Wang; data collection: Fengyue Jin; analysis and interpretation of results: Hongxin Yu, Yibing Wang, Meng Zhang, Anni Chen; draft manuscript preparation: Hongxin Yu, Yibing Wang. All authors reviewed the results and approved the final version of the manuscript.

**DECLARATION OF CONFLICTING INTERESTS**
The authors declared no potential conflicts of interest with respect to the research, authorship, and/or publication of this article.

**FUNDING**
This work was supported by the Provincial Key R&D Program of Zhejiang (2022C01129; 2024C01180), the National Natural Science Foundation of China (52272315), Ningbo International Science and Technology Cooperation Program (2023H020).

*Hongxin Yu, Yibing Wang, Fengyue Jin, Meng Zhang, and Anni Chen*

15. Li, Z., N. Kovachki, K. Azizzadenesheli, B. Liu, K. Bhattacharya, A. Stuart, and A. Anandkumar. Neural Operator: Graph Kernel Network for Partial Differential Equations. In ICLR 2020 Workshop on Integration of Deep Neural Models and Differential Equations, Addis Ababa, Ethiopia, 2020.

16. Lu, L., P. Jin, G. Pang, Z. Zhang, and G. E. Karniadakis. Learning Nonlinear Operators via DeepONet based On the Universal Approximation Theorem of Operators. *Nature machine intelligence*, Vol. 3, No. 3, 2021, pp.218-229.

17. Zhang, W., Y. Yu, Y. Qi, F. Shu, and Y. Wang. Short-term Traffic Flow Prediction based on Spatio-Temporal Analysis and CNN Deep Learning. *Transportmetrica A: Transport Science*, Vol. 15, No. 2, 2019, pp.1688-1711.

18. LeCun, Y., L. Bottou, Y. Bengio, and P. Haffner. Gradient-based Learning Applied to Document Recognition. *Proceedings of the IEEE*, Vol. 86, No. 11, 2002, pp.2278-2324.

19. Wang, S., and P. Perdikaris. Long-time Integration of Parametric Evolution Equations with Physics-informed Deeponets. *Journal of Computational Physics*, Vol. *475*, 2023, p.111855.

20. de Jong, T.O., K. Shukla, and M. Lazar. Deep Operator Neural Network Model Predictive Control. *arXiv preprint arXiv:2505.18008*, 2025.

21. Wang, Y., M. Papageorgiou, and A. Messmer. Real-Time Freeway traffic state estimation based on extended kalman filter: adaptive capabilities and real data testing. *Transportation Research Part A: Policy and Practice*, Vol. 42, 2008, pp. 1340-1358.

22. Wang, Y., M. Zhao, X. Yu, Y. Hu, P. Zheng, W. Hua, L. Zhang, S. Hu, and J. Guo. Real-time joint traffic state and model parameter estimation on freeways with fixed sensors and connected vehicles: state-of-the-art overview, methods, and case studies, *Transportation Research Part C: Emerging Technologies*, Vol. 134, 2022, Art. no. 103444.

23. Fan. S., and B. Seibold. Data-fitted First-order Traffic Models and Their Second-order Generalizations: Comparison by Trajectory and Sensor Data. *Transportation Research Part C: Emerging Technologies*, Vol. 2391, No. 1, 2013, pp. 32-43.

24. Wang, Y., M. Papageorgiou, and A. Messmer. RENAISSANCE-A unified macroscopic model-based approach to real-time freeway network traffic surveillance. *Transportation Research Part C: Emerging Technologies*, Vol.14, No. 3, 2006, pp. 190-212.

18